\documentclass{article}

\usepackage[preprint,sglblindworkshop]{Formatting_Instructions_For_NeurIPS_2026/neurips_2026}
\workshoptitle{Foundations of Agentic Systems Theory (FAST)}
\makeatletter
\renewcommand{\@noticestring}{Accepted at the Foundations of Agentic Systems Theory (FAST) Workshop at NeurIPS 2026.}
\makeatother

\usepackage[utf8]{inputenc}
\usepackage[T1]{fontenc}
\usepackage{hyperref}
\hypersetup{
  hidelinks,
  pdftitle={Before Agents Decide: Epistemic Action in LLM-Based Systems},
  pdfauthor={Yizhi Liu, Balaji Padmanabhan, and Siva Viswanathan}
}
\usepackage{url}
\usepackage{booktabs}
\usepackage{array}
\usepackage{microtype}
\usepackage{xcolor}
\usepackage{graphicx}
\usepackage{amsmath}

\title{Before Agents Decide: Epistemic Action in LLM-Based Systems}

\author{
Yizhi Liu$^{1}$ \qquad Balaji Padmanabhan$^{2}$ \qquad Siva Viswanathan$^{2}$\\[0.35em]
$^{1}$Fox School of Business, Temple University\\
$^{2}$Robert H. Smith School of Business, University of Maryland\\[0.25em]
\texttt{yizhi.liu@temple.edu} \quad \texttt{bpadmana@umd.edu} \quad \texttt{sviswan1@umd.edu}
}

\begin{document}

\maketitle
\vspace{-0.85em}

\begin{abstract}
Before a difficult decision, people often act simply to understand the situation better. We turn an object to see another side, place alternatives next to each other, or change one condition and observe what happens. These actions may not complete the task, but they improve the evidence needed for the next choice. LLM-based agents can search and explore, yet agent design gives less attention to an earlier question: is the available evidence ready for the decision? Sometimes necessary evidence is missing. In other cases, the evidence is present but its form hides what matters, or the comparison needed to judge it does not yet exist. Cognitive science calls actions that improve the basis for a later choice \emph{epistemic actions}. We bring this idea to LLM-based agents and distinguish three modes: acquiring missing evidence, transforming available evidence, and probing a system to create a revealing response. We use the term \emph{epistemic scaffolding} for the interfaces, tools, and environments that make these actions possible and auditable. This paper argues that agent design must address how decision-ready evidence is produced.
\end{abstract}

\section{Introduction}

Before a difficult decision, people often act simply to understand the situation better. We turn an unfamiliar object to inspect another side, place alternatives next to each other, or repeat a measurement under slightly different conditions. A Rubik's Cube solver may rotate the entire cube without moving any piece relative to another. The rotation does not solve the puzzle, but it reveals a hidden face and makes the next move easier to judge.

\citet{kirsh1994epistemic} call such behaviors \emph{epistemic actions}: actions that uncover information or reduce the cognitive work needed to use it. They contrast them with \emph{pragmatic actions}, which directly advance the task. One improves the basis for a choice; the other changes the world toward a goal. People move between the two, using one action to improve the evidence for the pragmatic action that follows.

Now place the same puzzle behind an interface that shows an agent only one face at a time. The agent may reason correctly about the visible face while the next move depends on stickers it cannot see. No amount of reasoning over that face will reveal them. Before choosing a move, the agent needs a way to change what it can observe.

LLM-based agents face the same limit whenever they must decide from what their tools and interfaces expose. Preparing evidence for the decision is rarely treated as a capability in its own right, so an agent may produce a polished answer even when its interaction never exposed what the decision requires.

We therefore argue that epistemic action should become a first-class part of agent design. Before answering or taking pragmatic action, an agent should judge whether its evidence is ready for the decision. If it is not, the agent may need to acquire, transform, or probe that evidence.

\begin{figure}[t]
  \centering
  \includegraphics[width=0.82\textwidth]{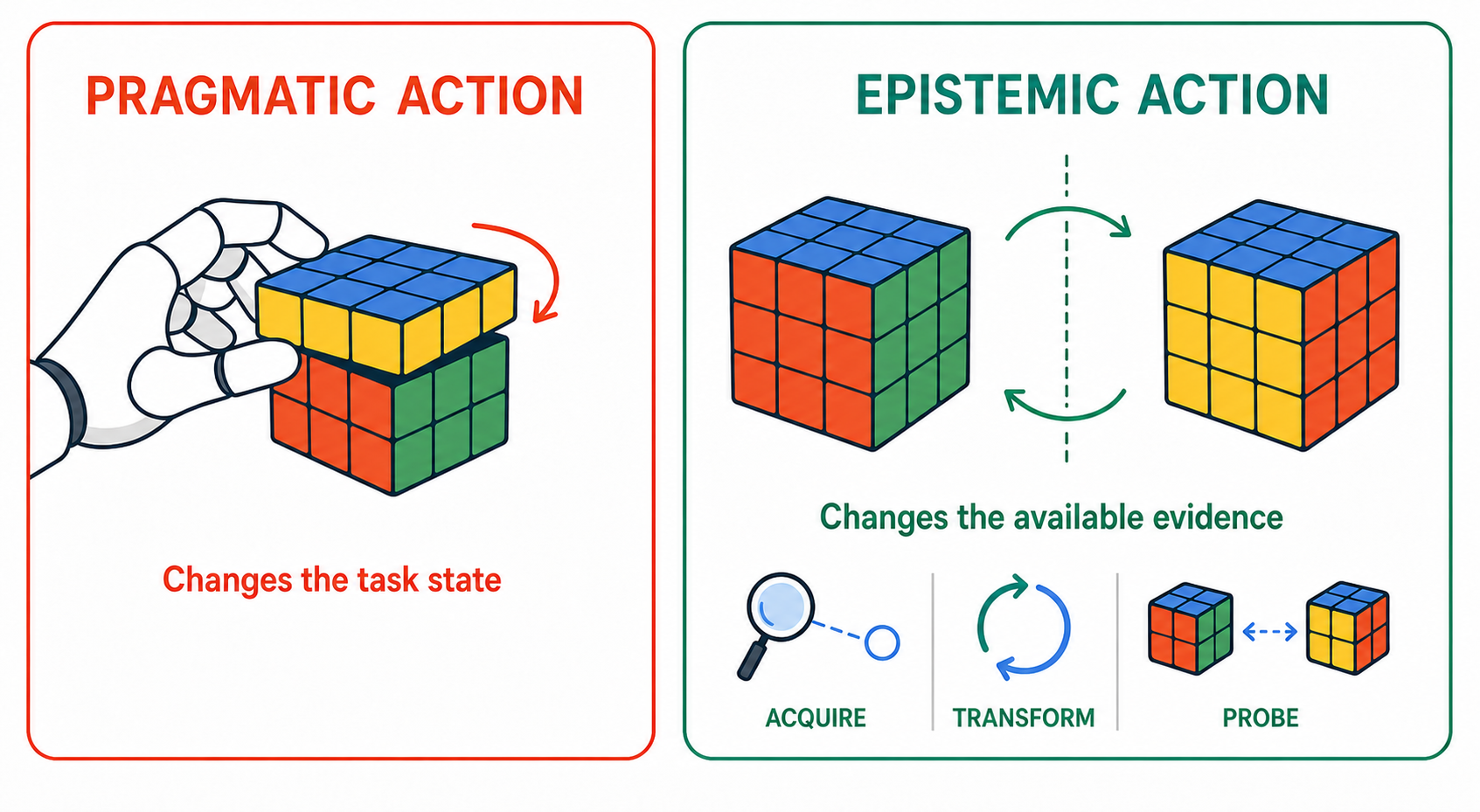}
  \vspace{-6pt}
  \caption{Pragmatic actions change the task state. Epistemic actions change the evidence for deciding what to do next. An agent may acquire, transform, or probe evidence before it commits.}
  \vspace{-10pt}
  \label{fig:epistemic}
\end{figure}

\section{Acting is not the same as acting for evidence}
\label{sec:distinction}

Turning one face of the cube changes the puzzle and may move the solver toward a solution. Rotating the whole cube leaves the puzzle unchanged but reveals another side. The first is pragmatic; the second prepares evidence for deciding what to do next.

LLM-based agents already search the web, inspect pages, edit files, and call tools \citep{yao2023react,yao2022webshop,yang2024sweagent}. Some of these actions serve an epistemic purpose. A WebShop agent, for example, may inspect another product because the purchase decision depends on a fact it has not yet seen. These systems already give agents some of the necessary tools. What remains less developed is the agent's own judgment of whether the available evidence is adequate and, if it is not, which action would improve it.

This capacity matters because reasoning can use only what the agent has observed. In the cube example, the colors on a hidden face cannot be inferred from the visible face alone. The same limit appears when an agent has not obtained a relevant fact, when the form of the evidence hides an important relationship, or when the answer depends on the result of an intervention that has never been tried. Longer reasoning leaves these problems unchanged. The agent must first act to obtain, reorganize, or produce the evidence that reasoning needs.

Cognitive science studies epistemic action; robotics studies active perception; and decision theory and partially observable planning ask when additional information is worth obtaining \citep{kirsh1994epistemic,bajcsy2018active,kaelbling1998planning}. Research in information systems shows how material arrangements and decision-support systems shape what a decision maker can notice and use \citep{orlikowski2006material,hesselmann2024scaffolding}. Research in strategy likewise shows how cognitive representations guide search and attention \citep{gavetti2000search}. Recent work describes agentic architectures as scaffolds for human strategic reasoning \citep{camuffo2026blackboxes}. These traditions share a simple lesson: what a decision maker can know depends partly on the actions available to them. We apply that lesson to LLM-based agents.

\section{Three modes of epistemic action}
\label{sec:modes}

Table~\ref{tab:modes} presents three modes of epistemic action that an agent should be able to consider before it commits: acquiring evidence it does not have, transforming evidence that is difficult to use, and probing for evidence that does not yet exist. A particular decision may require one mode, several modes, or none.

\clearpage
\begin{table}[t]
  \caption{Three modes of epistemic action.}
  \label{tab:modes}
  \centering
  \footnotesize
  \setlength{\tabcolsep}{3.5pt}
  \renewcommand{\arraystretch}{1.08}
  \begin{tabular}{@{}>{\raggedright\arraybackslash}p{1.75cm}>{\raggedright\arraybackslash}p{4.0cm}>{\raggedright\arraybackslash}p{7.1cm}@{}}
    \toprule
    Mode & Epistemic purpose & Examples \\
    \midrule
    Acquisitive & Obtain evidence that is not currently available & Ask a clarifying question; search; inspect another event; take a new measurement \\
    Transformative & Make available evidence more revealing or decision-appropriate & Change viewpoint; re-express a representation; decompose an aggregate; preserve structure during compression \\
    Probing & Create a diagnostic contrast or response & Construct a matched comparison; perturb one feature; run a counterfactual or controlled experiment \\
    \bottomrule
  \end{tabular}
\end{table}

\paragraph{Acquisitive actions.}
The most familiar problem is missing evidence. The agent can ask a clarifying question, retrieve a source, or inspect another part of its environment. In WebShop, agents reformulate queries, inspect product pages, and explore alternatives before choosing an item \citep{yao2022webshop}. ReAct similarly interleaves reasoning with retrieval from external sources \citep{yao2023react}. In each case, the epistemic action adds an observation before the agent chooses.

\paragraph{Transformative actions.}
Searching for more evidence is not always the answer. Sometimes the evidence is already available, but its form hides the structure that matters. It must then be transformed. People rotate objects, make notes, sort alternatives, and draw diagrams for this reason. SWE-agent restructures repository interaction and command feedback so that a language model can use them more effectively \citep{yang2024sweagent}. LLMLingua compresses long prompts while seeking to preserve their semantic content \citep{jiang2023llmlingua}. Designers choose both transformations in advance. An epistemically active agent would recognize when its current representation is inadequate and request or construct a better one.

\paragraph{Probing actions.}
Some evidence cannot be found in any existing source because it depends on a comparison or response that has not yet been produced. The agent must create a revealing situation and observe what happens. A controlled variation, for example, can show whether a judgment changes with one factor while the others remain stable.

In ScienceWorld, an agent can determine an unknown material's conductivity by assembling a circuit and observing the result \citep{wang2022scienceworld}. DiscoveryWorld asks agents to form hypotheses, design and run experiments, analyze the results, and act on their conclusions \citep{jansen2024discoveryworld}. \citet{zahavy2026jump} argues more broadly that scientific abduction requires more than passive prediction and points to action-controllable world models as a route to grounded simulation. In Zahavy's example, an agent would need to intervene in a simulated elevator by cutting the cable, creating an experience from which a new hypothesis might be formed.

Probing also applies when the required contrast cannot be observed directly. One study uses generative models to create matched patient images that vary perceived demographic presentation while preserving pain-related content \citep{liu2025deepfakes}. Its proof-of-concept multi-agent system generates, validates, and compares the variants in a bias audit. The original image contains no such comparison; the probe produces it for the question at hand.

When another agent or group is probed, the inquiry may alter the behavior it is meant to reveal. The resulting evidence must be interpreted together with the conditions that elicited it.

The modes can be combined: an agent may acquire a record, transform its representation, and probe what remains uncertain. It may also find that the current evidence already supports commitment. Epistemic competence includes knowing when another action is worth its cost.

\section{Epistemic scaffolding and a research agenda}
\label{sec:agenda}

Epistemic actions belong to the agent, while epistemic scaffolding belongs to the design around it. The term has been used in science education for support that shapes learners' knowledge practices \citep{lin2025epistemic}. Here it includes the interfaces, tools, representations, and environments that make epistemic actions possible. Static screenshots restrict acquisition, an irreversible summary limits transformation and checking, and a simulator without controllable variables prevents probing. The scaffold therefore helps set the boundary of what an agent can recognize, perform, and audit.

A useful scaffold records where new evidence came from, keeps transformations inspectable, and lets the agent stop when the likely gain no longer justifies the cost. These records also show whether a tool call improved the basis for the later decision.

Purpose is not visible in an action's surface form. An auditable trace must therefore record the pending decision, the evidence before and after the action, and the evidential change the agent expected; tool calls alone are insufficient.

The scaffold determines which epistemic actions the agent can consider. A standard value-of-information rule can rank those actions, but it cannot supply one that the scaffold omits. If the available set contains only search and retrieval, the agent cannot choose an action that re-expresses evidence or creates a diagnostic contrast, even with a perfect value estimate. We can state the familiar selection rule formally \citep{howard1966information}. Let $\mathcal{E}$ be the available epistemic actions, $I$ the current evidence, $I_e$ the evidence after $e$, $c(e)$ its cost, and $V_{\pi}(I)$ the expected value of the later pragmatic decision available to agent $\pi$:
\begin{equation}
e^{*}(I) \in \operatorname*{arg\,max}_{e\in\mathcal{E}\cup\{\emptyset\}}
\left\{\mathrm{E}\!\left[V_{\pi}(I_e)\mid I,e\right]-c(e)\right\}.
\label{eq:epistemic-value}
\end{equation}
The null action means committing now, with $I_{\emptyset}=I$ and $c(\emptyset)=0$. The decision rule is familiar; our claim concerns the construction of $\mathcal{E}$. The criterion is agent-relative, so a new representation may help $\pi$ use evidence that was already present. The scalar $c(e)$ is shorthand for multiple costs; privacy and safety may instead be constraints. Agent theory must explain how an agent estimates this value and validates the result.

Benchmarks that fix the evidence before the agent begins and score only the final answer cannot test this capacity. A minimal benchmark could hold the task, underlying model, prompt budget, and pragmatic actions fixed while varying whether acquisition, transformation, and probing are available. Tasks would require a missing observation, a change of representation, or a controlled intervention. Evaluation should separate three abilities: recognizing that the initial evidence is inadequate, selecting an epistemic action, and using its result. One comparison would give the agent the needed action explicitly; another would make the same action available without telling the agent when to use it. The difference would reveal whether the agent can identify its own epistemic need, rather than merely execute a prescribed step.

In multi-agent systems, one agent can become part of another's epistemic scaffold. A summary or handoff may omit provenance, conceal disagreement, or remove details that the receiving agent needs. Additional reasoning cannot reconstruct what never reached it. The receiver may instead need to request the source, compare alternative representations, or ask a targeted follow-up. Probing another agent raises a further issue because the question itself may change the behavior being observed. Evaluation should therefore track how evidence is produced and transformed across interactions, not only whether the group reaches the right answer.

Epistemic agency brings design tradeoffs that must remain visible. Transformations can discard information, probes can introduce artifacts, and additional observations can increase privacy or security exposure. Synthetic comparisons require careful validation because an intervention may change more than intended. These risks make auditability part of the framework: an evaluator should be able to inspect the evidence before and after an epistemic action and determine how it supported the eventual decision. The systems discussed above demonstrate individual capabilities. None yet chooses among all three modes and justifies both acting and stopping.

\section{Conclusion}

This paper began with a Rubik's Cube, where rotation can reveal the evidence needed for the next move without changing the puzzle state. LLM-based agents face the same design problem when the evidence required for a decision is missing, poorly represented, or not yet created. Acquisitive, transformative, and probing actions describe these ways of preparing evidence before commitment.

Epistemic scaffolding gives agents the means to take such actions and makes the resulting evidence auditable. This paper therefore places the preparation of evidence alongside reasoning as a central problem in the design of reliable agents.

\clearpage
\setlength{\bibsep}{3pt plus 0.3ex}
\bibliographystyle{plainnat}
\bibliography{references_epistemic}

\end{document}